%% file: main.tex
\documentclass[reprint, amsmath,amssymb, aps, pra]{revtex4-2}
\input{settings/packages.tex}
\hypersetup{breaklinks=true}

\begin{document}

\preprint{APS/123-QED}
\title{
    \texorpdfstring{
        A Machine Learning Approach That Beats Large Rubik's Cubes
        \\ \small{The CayleyPy Project}
    }{
        A Machine Learning Approach That Beats Large Rubik's Cubes
    }
}

\input{settings/autors.tex}

\input{settings/abstract.tex}

\maketitle

\section{\label{intro}Introduction}
\input{parts/intro.tex}

\section{\label{sec_results}Results}
\input{parts/results-I.tex}
\input{parts/results-II.tex}

\input{parts/tab1.tex}
\input{parts/results-III.tex}

\section{\label{sec_discussion}Discussion}
\input{parts/discussion.tex}

\section{\label{sec_methods} Methods}
\input{parts/methods-I.tex}

\input{parts/methods-II.tex}

\input{parts/tab2.tex}
\input{parts/methods-III.tex}
\input{parts/methods-IV.tex}
\input{parts/tab3.tex}
\input{parts/methods-V.tex}

\section{Acknowledgments}
\input{parts/acknowledgments.tex}

\input{parts/availability.tex}

\bibliography{bibliography}

\end{document}

%% file: settings/packages.tex
\usepackage{graphicx}
\usepackage{dcolumn}
\usepackage{bm}

\usepackage{orcidlink}
\usepackage{breakurl}
\usepackage{hyperref}
\usepackage{booktabs}

%% file: settings/autors.tex
\author{
	A. Chervov\orcidlink{0000-0003-4564-3272}$^{1,2,3}$\textsuperscript{†}, 
	K. Khoruzhii\orcidlink{0000-0003-4689-3812}$^4$\textsuperscript{†}, 
	N. Bukhal$^5$, 
	J. Naghiyev$^1$, 
	V. Zamkovoy$^5$, 
	I. Koltsov\orcidlink{0009-0009-3553-9069}$^5$, 
	L. Cheldieva$^5$, 
	A. Sychev\orcidlink{0009-0008-2455-0310}$^5$, 
	A. Lenin\orcidlink{0009-0002-2102-483X}$^5$, 
	M. Obozov$^6$, 
	E. Urvanov$^5$, 
	A. Romanov\orcidlink{0000-0003-1838-0489}$^5$\textsuperscript{*}
}
\affiliation{
	$^1$Institut Curie, Université PSL, Paris, F-75005, France;  \\
	$^2$INSERM U900, Paris, F-75005, France;  \\
	\mbox{$^3$CBIO, Mines ParisTech, Université PSL, Paris, F-75005, France;}  \\
	\mbox{$^4$Technical University of Munich, Garching, 85748, Germany;}  \\
	\mbox{$^5$Institute of Artificial Intelligence, RTU MIREA, Moscow, 119454, Russia;} \\
	$^6$Yandex Research, Moscow, 119021, Russia
}

\author{ \small
	\textsuperscript{*}Corresponding author: \href{mailto:romanov@mirea.ru}{romanov@mirea.ru} \\
	\textsuperscript{†}These authors contributed equally to this work.
}
\noaffiliation

%% file: settings/abstract.tex
\begin{abstract}
The paper proposes a novel machine learning-based approach to the pathfinding problem on extremely large graphs. This method leverages diffusion distance estimation via a neural network and uses beam search for pathfinding. We demonstrate its efficiency by finding solutions for 4$\times$4$\times$4 and 5$\times$5$\times$5 Rubik's cubes with unprecedentedly short solution lengths, outperforming all available solvers and introducing the first machine learning solver beyond the 3$\times$3$\times$3 case. In particular, it surpasses every single case of the combined best results in the Kaggle Santa 2023 challenge, which involved over 1,000 teams. For the 3$\times$3$\times$3 Rubik's cube, our approach achieves an optimality rate exceeding 98\%, matching the performance of task-specific solvers and significantly outperforming prior solutions such as DeepCubeA (60.3\%) and EfficientCube (69.6\%). Additionally, our solution is more than 26 times faster in solving 3$\times$3$\times$3 Rubik's cubes while requiring up to 18.5 times less model training time than the most efficient state-of-the-art competitor.
\end{abstract}

%% file: parts/intro.tex
Rubik's cube is one of the most famous puzzles, which is believed to be played by more than a billion people in the world~\cite{rubik2020cubed}. According to~\cite{van2002inventing}, it was included in the 100 most influential inventions of the 20th century. Even decades after its first introduction, it is still used as a benchmark and model task in various fields: 
artificial intelligence~\cite{agostinelli2019solving},
robotics~\cite{openai2019solvingrubikscuberobot},  
graphs algorithms~\cite{korf2008linear, sturtevant2013minimizing}, 
cryptography \cite{petit2013rubik},
image encryption \cite{loukhaoukha2012secure}, 
statistical physics~\cite{chen2014rubik, gower2024saddles}, 
group theory~\cite{joyner2008adventures, cornock2015teaching}, 
for human cognitive abilities~\cite{meinz2023ability}.

From a more general perspective, solving the Rubik's Cube is a particular case of a planning problem---one needs to plan a sequence of actions to transit between two given states. Planning robot moves and games like chess or Go represent similar problems---for example, a game's goal is to plan moves to transit from the initial position to the winning position(s). The mathematical framework for such problems is pathfinding on graphs (state transition graphs): all possible states are represented as nodes, and edges correspond to transitions between states based on actions (moves). The planning task thus reduces to finding a path from a given initial node to one or more desired nodes. A specific class of graphs represents the Rubik's Cube and similar puzzles---Cayley-type graphs of the puzzle's symmetry group. These are highly symmetric state transition graphs where the symmetry group can transform any node into another.
Cayley graphs are of fundamental importance in modern mathematics~\cite{gromov1993geometric, tao2015expansion} and have numerous applications: in bioinformatics for estimating evolutionary distances~\cite{Pevzner1995human2mice, Pevzner1999cabbage2turnip, wilson2024cayley, bulteau2019parameterized}; in processor interconnection networks~\cite{akers1989group,cooperman1991applications,heydemann1997cayley}; in coding theory for the construction of expander graphs and related codes~\cite{hoory2006expander}; in cryptography for constructing specific hash functions~\cite{zemor1994hash,petit2013rubik}; in machine learning~(ML)~\cite{wilson2024cayley}; and in quantum computing~\cite{ruiz2024quantum,sarkar2024quantum,dinur2023good, acevedo2006exploring, gromada2022some}. 

Finding the shortest paths on generic finite Cayley graphs is an NP-hard problem~\cite{even1981minimum}, as it is for many particular groups: the Rubik's Cube group~\cite{demaine2017solving} and some others~\cite{bulteau2015pancake, wilson2024cayley}.
Brute force breadth-first search, Dijkstra's, and related methods can find the shortest paths on graphs with billions of nodes, the bidirectional trick squares feasible sizes, but these methods require extremely large computational resources and are not practical for much larger sizes, which are of our interest.
Moreover, no effective tools are currently available to find any (not just the shortest) paths on Cayley graphs of large finite groups. For example, modern computer algebra systems like GAP~\cite{linton2007gap} fail on any sufficiently large group, such as the 4$\times$4$\times$4 Rubik's Cube.

{\bf Results.} To address these issues, we develop machine learning-based methods to find paths on a broad class of graphs (specified below) of unprecedented sizes and the ability to produce unprecedentedly short paths. In the present paper, we provide code applicable to Cayley graphs (or, more generally, Schreier graphs) of any finite permutation group and focus on demonstrating its efficiency for Rubik's groups.
The presented approach is the first machine learning-based method to successfully solve the 4$\times$4$\times$4 and 5$\times$5$\times$5 Rubik's Cubes, with  $7.4\times{10}^{45}$ and $ 1.2\times{10}^{74}$ elements, respectively. The obtained solution lengths are shorter than those produced by any available method, including the combination of top results from the Kaggle Santa 2023 Challenge~\cite{santa-2023}, where more than a thousand participants applied and developed various methods.
Moreover, for the 4$\times$4$\times$4 cube, the average solution length is 46.51, which is below the conjectural diameter 48~\cite{hirata2024graph}, thus providing further evidence for the quality of our solutions.
For the 3$\times$3$\times$3 Rubik's Cube, we achieve 98.4\% optimality solving scrambles from the DeepCubeA dataset, surpassing previous machine learning approaches: DeepCubeA at 60. 3\%~\cite{agostinelli2019solving} and EfficientCube at 69.6\%~\cite{takano2023selfsupervision}. 

To conclude, this research aims to make a significant step in advancing machine learning applications to graph pathfinding and to demonstrate its efficiency in the case of Rubik's cubes of different sizes, providing more optimal solutions than any available approach for large groups.
The main contributions are the following:

\begin{enumerate}
    \item We propose a novel multi-agent, machine learning-based approach to find paths on Cayley graphs of finite groups. It is the first machine learning approach capable of handling groups as large as $ {10}^{74}$. It achieves over 98\% optimality on the DeepCubeA dataset of 3$\times$3$\times$3 cubes, reaching the level of task-oriented solvers based on pattern databases. It produces better results (shorter solution paths) than any known competitor for 4$\times$4$\times$4 and 5$\times$5$\times$5 Rubik's Cubes, including the aggregated best results from the 2023 Kaggle Santa Challenge, representing the current state of the art.
    \item We demonstrate that increasing the size of the set used to train multilayer perceptrons with residual blocks has a limited impact on the pathfinder's performance. At the same time, increasing the beam width and number of agents robustly improves the average solution length and optimality. This surprising finding helped choose the size of the train data for each agent and achieve best-in-class performance without wasting computational resources on additional training.
    \item The training time and computational resources required for our approach are significantly smaller than those for state-of-the-art approaches. Our solution, tested on the same hardware and beam width as EfficientCube (the previous leading ML solution), performs pathfinding slightly better than EfficientCube, solving the task $\approx26$x faster and requiring up to 18.5x less model training time than the competitor.
\end{enumerate}

In recent years, machine learning has been emerging as "a tool in theoretical science"~\cite{douglas2022machine}, leading to several noteworthy applications to mathematical problems~\cite{lample2019deep,davies2021advancing, bao2021polytopes, romera2024mathematical, coates2024machine,alfarano2024global, charton2024patternboost,shehper2024makes}. This research is part of the larger project, which aims to create an open-source machine learning Python framework for analyzing Cayley graphs and contribute to the fascinating, emerging area of machine learning applications in theoretical sciences.

%% file: parts/results-I.tex
\subsection{\label{sec_approach} Proposed Machine Learning Approach}

This paper presents a unified approach for finding paths on a large class of graphs, focusing on demonstrating its efficiency for Rubik's cube graphs. It does not rely on any prior knowledge or human expertise about the graphs.
The approach has two main components: a neural network model and a graph search algorithm -- similar to previous works such as AlphaGo/AlphaZero~\cite{silver2016mastering, silver2017mastering}, DeepCube~\cite{mcaleer2019solving, agostinelli2019solving}, and EfficientCube~\cite{takano2023selfsupervision}, among others. The model is trained to guide what moves should be done to get closer to the destination node ("solved state" for puzzles). The graph search algorithm starts from a given node and moves to nodes closer to the destination, based on the neural network's predictions, until the destination node is found.

The basic assumption on a graph is that there is a vector associated with each node (feature vector). These vectors serve as an input for the neural network \footnote{The precise quantification of requirements for feature vectors that would ensure the successful operation of the proposed method is challenging. We aim to demonstrate its efficiency in the context of Rubik’s group cases. On one extreme, even random vectors suffice if the training data covers all nodes — an idea employed in well-known approaches such as DeepWalk~\cite{perozzi2014deepwalk} and Node2vec~\cite{grover2016node2vec}. However, our focus is different: only a small subset of nodes will be covered by the training data (random walks). The key point is the ability of the neural network to generalize from that small subset to the entire graph — something that is impossible with random features. Worse, the feature vectors are related to the distance between nodes on a graph --- more training data is required, and more advanced parameters and resources should be used at all steps of the proposed method. The role of the neural network is to transform the initial feature vectors into a latent representation, where nodes that are closer on the graph are also closer in the latent space. }. For puzzles or permutation groups---it is just the vector describing the permutation $p$ of $l$-symbols, i.e. vector of numbers $(p(0),p(1),\ldots,p(l-2),p(l-1))$. Additionally, we assume that a specific node on the graph, such as the 'solved state' for puzzles, is selected. The task is to find a path from any given node to this selected node. Since the graph sizes may exceed $10^{70}$, standard pathfinding methods are not applicable. 

\begin{figure*}
\centering
\includegraphics[width=1\textwidth]{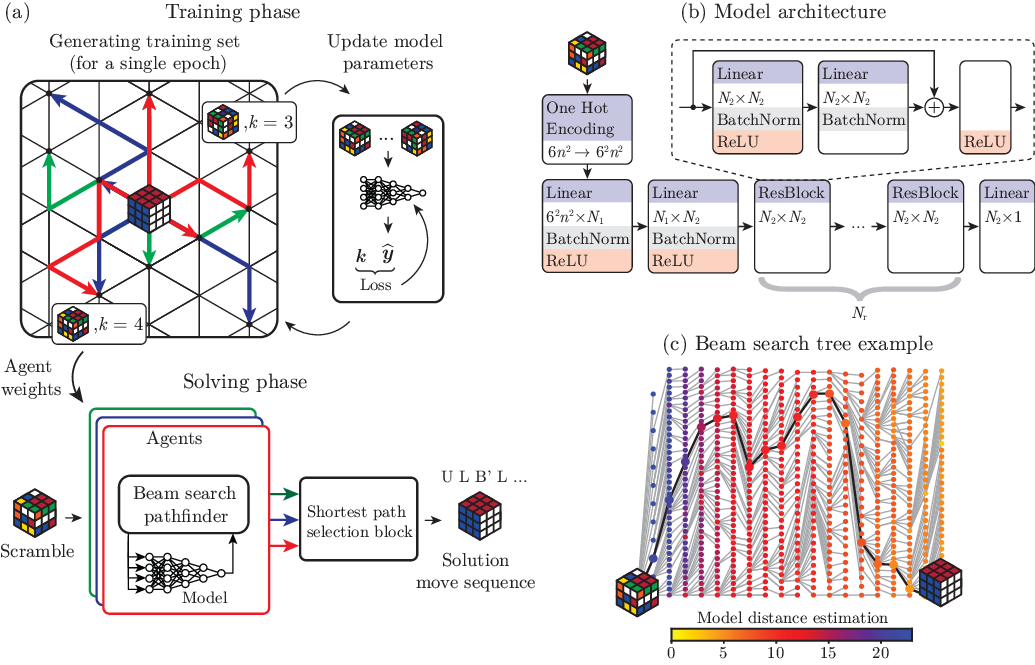}
\caption{Proposed ML solution for Rubik's cube solving: (a)~proposed multi-agent solver's process flow; (c)~ResMLP neural network architecture; (b)~an example of beam search pathfinding on 3$\times$3$\times$3 cube's graph using $W=40$.}
\label{fig_solution_struct}
\end{figure*}

The key steps of the proposed method are illustrated in the figure \ref{fig_solution_struct}a and described below:

 {\bf  Generating Training Data via Random Walks and Diffusion Distance.} Generate $N$ random walk trajectories starting from a selected node. (The generation of a random walk is a simple process: select a random neighbor of the current node and repeat this process iteratively for multiple steps.) Each random walk trajectory consists of up to $K_{\text{max}}$ steps, where $N$ and $K_{\text{max}}$ are integer parameters of the method. For some nodes encountered during the random walks, we store a set of pairs $(v, k)$, where $v$ represents the vector corresponding to the node and $k$ is the number of steps required to reach it via the random walk. This set will serve as the training data. For the Rubik's Cube, random walks correspond to random scrambling: starting from the "solved state," we perform a series of random scrambles and record the resulting positions and the number of scrambles performed. Conceptually, in the limit as $N \to \infty$, the average value of $k$ measures the "diffusion distance"---roughly speaking, the length of the random path or an estimate of how quickly diffusion reaches a given node. In contrast to the DAVI approach used in \cite{agostinelli2019solving}, random walk generation is very computationally cheap, making it possible to generate them directly during the training procedure.

 {\bf Training the neural network.} The generated set of pairs $(v, k)$ serves as the training set for the neural network. Specifically, $v$ serves as the 'feature vector' (the input for the neural network), and $k$ represents the 'target' (the output the network needs to predict). Thus, the neural network's predictions for a given node $v$ estimate the diffusion distance from $v$ to the selected destination node (solved state of the puzzle). We utilize a multilayer perceptron~(MLP) architecture with several residual blocks and batch normalization, as shown in Figure~\ref{fig_solution_struct}b, which will be further called ResMLP. It is a general form of the MLPs used in \cite{agostinelli2019solving,takano2023selfsupervision}. All the models are trained in advance before the solving phase.

 {\bf Pathfinding with neural network heuristics and Beam Search.} This step finds a path from a given node to the destination node. The neural network provides heuristics on where to make the next steps, while the graph pathfinding technique compensates for any possible incorrectness in the neural network predictions. The beam search pathfinding method is quite simple but has proven to be the most effective for us and works as follows. Fix a positive integer $W$ — a parameter known as the "beam width" (or "beam size"). Starting from a given node, we take all its neighboring nodes and compute the neural network predictions for all of them. We then select the $W$ nodes closest to the destination according to the neural network (i.e., the predictions have smaller values). We take these selected $W$ nodes' neighbors, drop duplicates, and again compute the neural network predictions, choosing the top $W$ nodes with the best (i.e., minimal) predictions. The search iterations are repeated until the destination node is found (or the limit of steps is exceeded). The whole process is illustrated in Figure~\ref{fig_solution_struct}c.
  
 {\bf Multi-agency.} The method described in the steps above relies on random walks for train set creation, and thus, due to that randomness, each new launch will create a new train set, and thus, each new neural network approximates the distance differently. This diversity is large enough to yield a new solution path for each launch typically. And hence, typically, several repetitions allow for the discovery of a shorter path than a single run. We call each trained neural network an agent. To solve any given state - we solve it with all the agents and then choose the best result (the shortest solution path among all the agents) -- illustrated in Figure~\ref{fig_solution_struct}a. 

%% file: parts/results-II.tex
\subsection{\label{sec_analysis} Optimality vs the Proposed Approach Parameters }

\begin{figure*}[ht]
\centering
\includegraphics[width=\textwidth]{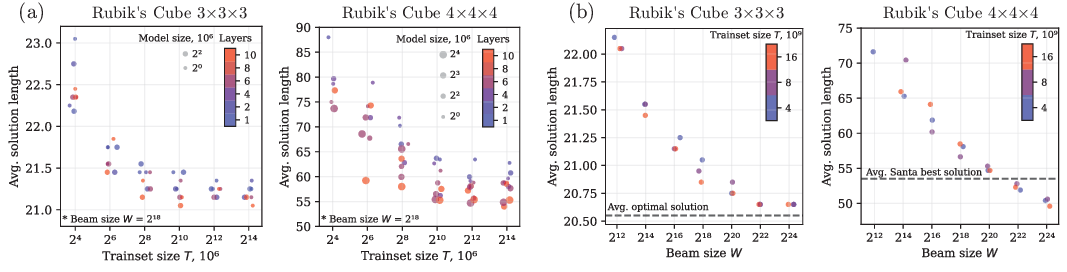}
\caption{Influence of model parameters on solution length for 3$\times$3$\times$3 and 4$\times$4$\times$4 cubes (jitter plot): (a)~influence of the model size, trainset sizes, and model depth on average solution length; (b)~influence of the beam width on average solution length.}\label{fig_param_vs_len}
\end{figure*}

The proposed solver has the following main parameters $A$ -- the number of agents, $W$ -- beam width used by each agent during pathfinding and ResMLP model general parameters: $N_1$ -- the size of the first layer, $N_2$ -- the size of the second layer and residual blocks' layers, $N_r$ -- the number of residual blocks and $T$ -- the trainset size. For easier comparison, $N_1$, $N_2$, and $N_r$ are also summarized by model size $P$ -- the total number of ResMLP parameters~(weights and biases). In this section, we analyze the influence of these parameters on the solver's average solution length and optimality.

{\bf Train set size.} First, in the example of 3$\times$3$\times$3 and 4$\times$4$\times$4 Rubik's cubes, we analyzed how the model and trainset sizes, as well as model depth, influence the average solution length using a single agent with fixed beam width ($W=2^{18}$). The experiment details are provided in Section~\ref{sec_exp_design}, while the results are presented in Figure~\ref{fig_param_vs_len}a. It is seen from Figure~\ref{fig_param_vs_len}a that from a certain point, the raise of $T$ does not lead to any significant reduction of average solution length, especially considering the fact that the trainset size is demonstrated in logarithmic scale. Even more surprising, the $T$ value corresponding to this point is very similar for 3$\times$3$\times$3 and 4$\times$4$\times$4 cubes and neural networks of different sizes and depths. Thus, the experiments above reveal a rather unexpected effect - performance stagnation with respect to the train size.  

{\bf MLP layers and sizes.} As expected, larger and deeper networks trained on train sets of the same size generally provide shorter solutions than smaller models. What is less expected is that the higher number of layers~(higher $N_1$, $N_2$, and $N_r$) is more significant than a larger number of parameters $P$. More surprising is that even small models with 1M of parameters can reach the average solution length comparable to DeepCubeA and EfficientCube using neural networks with $\approx14.7\mbox{M}$ parameters. Based on these observations for further consideration, we used a deep neural network having the same number of layers~(ten) as \cite{agostinelli2019solving} and \cite{takano2023selfsupervision}, but with the smaller model size of 4M parameters.

{\bf Beam width.} It is the most important parameter. We performed multiple tests on a single agent equipped with this model, changing $W$ from $2^{12}$ to $2^{24}$. The results of these tests are presented in Figure~\ref{fig_param_vs_len}b~(the exact model parameters and details of the tests are provided in Section~\ref{sec_exp_design}). From Figure~\ref{fig_param_vs_len}b, it is clear that increasing $W$ effectively reduces the average solution length. Moreover, the solution length decreases approximately linearly with the logarithm of the beam width  $W$. On the 3$\times$3$\times$3 cube, increasing $W$ up to $2^{24}$ allows us to get close to the optimal solution, while for 4$\times$4$\times$4, the same beam width results in a better average solution length than the best ones submitted to the 2023 Santa Challenge. 

\begin{figure*}[ht]
\centering
\includegraphics[width=\textwidth]{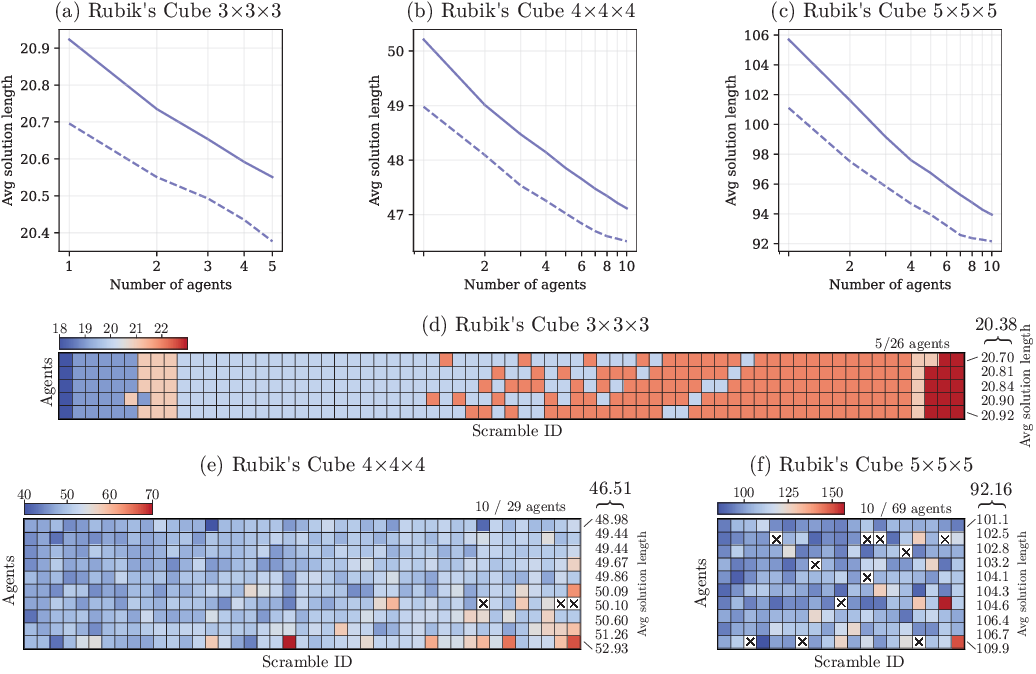}
\caption{ Average solution length of the proposed multi-agent approach depending on the number of agents composing its output for (c)~3$\times$3$\times$3, (e)~4$\times$4$\times$4, and (f)~5$\times$5$\times$5 Rubik's cubes. Solid line -- random set of the agents, dashed line -- best set. Distribution of solution lengths for (c)~3$\times$3$\times$3, (d)~4$\times$4$\times$4, and (e)~5$\times$5$\times$5 Rubik's cubes for the best ensemble. }\label{fig_multi_agent}
\end{figure*}

{\bf Agents number.} In the third part of the experimental studies, we investigated the influence of the number of agents $A$ on the solver's efficiency. These experiments were performed on 3$\times$3$\times$3, 4$\times$4$\times$4, and 5$\times$5$\times$5 Rubik's Cube. We used 10-layer ResMLP models with 4M parameters trained on 8B states in all the cases. The beam width was chosen $W=2^{24}$ so each agent could fit into the memory of a single GPU regardless of the solved cube size. The details of the performed experiments are available in Section~\ref{sec_exp_design}, while their results are provided in Figure~\ref{fig_multi_agent}. For ease of analysis, Figures~\ref{fig_multi_agent}a,\ref{fig_multi_agent}b,\ref{fig_multi_agent}c demonstrate lengths only for those agents whose solutions at least once were used as the solver's output.

Figure~\ref{fig_multi_agent} clearly shows that the average solution rate of a multi-agent is always higher than the one achieved by the best single agent~(up to 8 moves for the 5$\times$5$\times$5 cube). Solid lines on Figures~\ref{fig_multi_agent}a, \ref{fig_multi_agent}b, \ref{fig_multi_agent}c show how the size of the ensemble influences the average solution length for the random set of the agents. As seen in all three cases of 3$\times$3$\times$3, 4$\times$4$\times$4, and 5$\times$5$\times$5 Rubik's cubes, the larger number of agents robustly provided more optimal pathfinding. The dashed line demonstrates the same dependency but for the set of agents jointly providing the best overall solution. Figures~\ref{fig_multi_agent}c, \ref{fig_multi_agent}d, \ref{fig_multi_agent}e demonstrate in color code how each agent from this set participates in the final solution for every scramble from the dataset. The scrambles which were not solved in less than 200 moves are marked with crosses.

If it is seen from Figures~\ref{fig_multi_agent}c,\ref{fig_multi_agent}d,\ref{fig_multi_agent}e that the worst agents in the ensemble not only provide much longer results than the final solution but also include multiple scrambles that were unsolved. In the case of using the single-model approach, these agents would be considered unsatisfyingly trained. Nevertheless, they are included in the best ensemble because they provided the shortest solution on one or two scrambles. Moreover, our approach reached efficiency unreachable for other ML solutions only due to such specialized agents.

Even though the results presented in Figure~\ref{fig_multi_agent} on 3$\times$3$\times$3, 4$\times$4$\times$4, and 5$\times$5$\times$5 cubes can be achieved using 5, 10, and 10 agents, respectively, the probability of training all these agents in a row is very low. For example, to beat all the 5$\times$5$\times$5 scrambles from the 2023 Santa Challenge dataset, we trained 69 different agents, while further analysis showed that only 10 of them composed all the output results. At the same time, the first agent trained to solve 4$\times$4$\times$4 cubes beat all the respective scrambles from the mentioned dataset but did not even get in the final ensemble because multiple other agents jointly surpassed it. Thus, achieving a high level of optimality requires many agents, as seen from the logarithmic nature of the plots demonstrated in Figure~\ref{fig_multi_agent}. Nevertheless, due to the high scalability of the proposed approach and the ability to run on distributed hardware using dozens of independent agents, it is not an issue using modern computational hardware.

%% file: parts/tab1.tex
\begin{table*}
\caption{The most notable results achieved by the proposed solution and comparison with competitors.}\label{tbl_best}
\begin{tabular*}{\textwidth}
{@{}>{\raggedright\arraybackslash}p{0.7cm} 
>{\raggedright\arraybackslash}p{5.2cm} 
>{\raggedright\arraybackslash}p{3cm} 
>{}p{0.7cm}
>{}p{0.7cm}
>{}p{1cm}
>{}p{0.9cm}
>{}p{2cm} 
>{}p{2cm}@{}}
\toprule
\textbf{No.} & \textbf{Solver} & \textbf{Metric\footnotemark[1], Size, Dataset} & \multicolumn{4}{c}{\textbf{Solver parameters}} & \textbf{Average solution length} & \textbf{Optimality/ Superiority}  \\
\noalign{\vspace{-0.5cm}}
\cmidrule(lr){4-7}
& & & \textit{\textbf{A}} & \textit{\textbf{W}} & \textit{\textbf{P}} & \textit{\textbf{T}} & & \\
\midrule
\multicolumn{9}{c}{ \textbf{2x2x2 Rubik's cube}}\\
\midrule
1 & Genetic~\cite{swita2023solving}  &  HTM, 1, \cite{swita2023solving} & n/a & n/a & n/a & n/a & 30\footnotemark[2]  & n/a\footnotemark[2] \\
2 & Breadth First Search  & QTM, 100, Ours & n/a & n/a & n/a & n/a & 10.669  & Opt. 100\%\\
\textbf{3} & Ours, 1-layer MLP  & QTM, 100, Ours & 1 &$2^{18}$ & 0.15M & 8B & 10.669  & Opt. 100\% \\
\textbf{4} & Ours, 10-layer ResMLP  & QTM, 100, Ours & 1 & $2^{18}$ & 0.92M & 8B & 10.669  & Opt. 100\% \\
\midrule
\multicolumn{9}{c}{ \textbf{3$\times$3$\times$3 Rubik's cube}}\\
\midrule
5 & Genetic~\cite{swita2023solving}  &  HTM, 1, \cite{swita2023solving} & n/a & n/a & n/a & n/a & 238\footnotemark[2]  & n/a\footnotemark[2] \\
6 & Optimal PDB+\footnotemark[3] solver~\cite{agostinelli2019solving} & QTM, 1000, \cite{agostinelli2019solving} & n/a & n/a & n/a & n/a & 20.637 & Opt. 100\%\\
7 & DeepCube~\cite{agostinelli2019solving}  & QTM, 1000, \cite{agostinelli2019solving} & 1 & n/a & 14.7M & 10B  & 21.50  & Opt. 60.3\%\\
8 & EfficientCube~\cite{takano2023selfsupervision}  & QTM, 1000, \cite{agostinelli2019solving} & 1 & $2^{18}$ & 14.7M & 52B   & 21.26  & Opt. 69.6\%\\
9 & EfficientCube~\cite{takano2023selfsupervision} (reproduced)  & QTM, 1000, \cite{agostinelli2019solving} & 1 & $2^{18}$ & 14.7M & 52B   & 21.255  & Opt. 69.8\%\\
\textbf{10} & \textbf{Ours}, 10-layer ResMLP  & QTM, 1000, \cite{agostinelli2019solving} & 1 & $2^{18}$ & 4M & 8B & \textbf{21.137} & Opt. \textbf{75.4}\% \\
\textbf{11} & \textbf{Ours}, 1-layer MLP  & QTM, 1000, \cite{agostinelli2019solving} & 1 &  $2^{24}$ & 0.34M & 8B & \textbf{20.829} & Opt. \textbf{90.4}\% \\
\textbf{12} & \textbf{Ours}, 10-layer ResMLP  & QTM, 1000, \cite{agostinelli2019solving} & 1 & $2^{24}$ & 4M & 8B & \textbf{20.691} & Opt. \textbf{97.3}\% \\
\textbf{13} & \textbf{Ours}, multi-agent 10-layer ResMLP  & QTM, 1000, \cite{agostinelli2019solving} & 26 & $2^{24}$ & 4M & 8B & \textbf{20.669} & Opt. \textbf{98.4}\% \\
14 & Santa Challenge~\cite{santa-2023} & UQTM, 82, \cite{santa-2023} & n/a & n/a & n/a & n/a & 21.829 & n/a\footnotemark[4] \\
\textbf{15} & \textbf{Ours}, 10-layer ResMLP  & UQTM, 82, \cite{santa-2023} & 1 & $2^{24}$ & 4M & 8B & \textbf{19.512 } & Sup. \textbf{100}\% \\
\midrule
\multicolumn{9}{c}{ \textbf{4$\times$4$\times$4 Rubik's cube}}\\
\midrule
16 & Genetic~\cite{swita2023solving}  &  HTM, 1, \cite{swita2023solving} &  n/a & n/a & n/a & n/a & 737\footnotemark[2]  & n/a\footnotemark[2] \\
17 & Santa Challenge~\cite{santa-2023} & UQTM, 43, \cite{santa-2023} &  n/a & n/a & n/a & n/a & 53.49  & n/a\footnotemark[4] \\
\textbf{18} & \textbf{Ours}, 10-layer ResMLP  & UQTM, 43, \cite{santa-2023} & 1 & $2^{24}$ & 4M & 8B & \textbf{48.98}  & Sup. \textbf{100}\% \\
\textbf{19} & \textbf{Ours}, multi-agent 10-layer ResMLP  & UQTM, 43, \cite{santa-2023} & 29 &  $2^{24}$ & 4M & 8B & \textbf{46.51}  & Sup. \textbf{100}\% \\
\midrule
\multicolumn{9}{c}{ \textbf{5$\times$5$\times$5 Rubik's cube}}\\
\midrule
20 & Genetic~\cite{swita2023solving}  &  HTM, 1, \cite{swita2023solving} & n/a & n/a & n/a & n/a & 1761\footnotemark[2]  &  n/a\footnotemark[2]\\
21 & Santa Challenge~\cite{santa-2023} & UQTM, 19, \cite{santa-2023} & n/a & n/a & n/a & n/a & 96.58  & n/a\footnotemark[4] \\
\textbf{22} & \textbf{Ours}, multi-agent 10-layer ResMLP  & UQTM, 19, \cite{santa-2023} & 69 & $2^{24}$ & 4M & 8B & \textbf{92.16}  & Sup. \textbf{100}\% \\
\midrule
\end{tabular*}

\footnotetext[1]{HTM -- half-turn metric, QTM -- quater-turn metric. The 2023 Kaggle Santa Challange dataset uses modified QTM with unfixed corners and centers of the cube, which is marked UQTM.}
\footnotetext[2]{The results presented in \cite{swita2023solving} evaluated using single cube scrambled with 100 random moves. For these results, column "Avg. solution len" contains the minimal length achieved by the most suitable genetic algorithm configuration. The optimality/superiority was not evaluated for these results.}
\footnotetext[3]{\url{https://github.com/rokicki/cube20src}}
\footnotetext[4]{Superiority is not applicable for 2023 Kaggle Santa Challenge best results as \cite{santa-2023} dataset is built of these results.}
\end{table*}

%% file: parts/results-III.tex
\subsection{\label{sec_best} Results Summary and \texorpdfstring{\\}{} Comparison with Prior Art}

Table~\ref{tbl_best}\footnote{All the solvers presented in the Table~\ref{tbl_best} managed to solve all the scrambles from the listed datasets.} summarizes the main results achieved by the proposed solver, highlighting its Superiority over the prior state of the art. Notably, it surpasses the 2023 Kaggle Santa Challenge results, where over a thousand teams competed in virtual puzzle solutions, representing the best available methods and results. It should be mentioned that we were limited in computation resources during our research. Thus, our results can be improved even more by using more advanced hardware, which will allow for an increase in beam width and the number of agents.

A single agent with single-layer MLP can solve all the DeepCubeA dataset with 90.4\% of optimality, significantly enhancing results of the most advanced state-of-the-art ML solutions: DeepCubeA and EfficientCube. 26 agents equipped with 10-layer ResMLP models managed to solve all 1000 scrambles from DeepCubeA dataset with 97.6\% optimality, which is the best result ever achieved by any ML solution (significantly surpassing 60.3\%, 69.8\% results from DeepCubeA and EfficientCube). A single-agent solution implemented using our approach and 10-layer ResMLP managed to beat each best result corresponding to 3$\times$3$\times$3 and 4$\times$4$\times$4 Rubik's cubes submitted on the 2023 Kaggle Santa Challenge (averages: 48.98 vs 53.49). At the same time, 29 agents managed to solve all the 4$\times$4$\times$4 cube's scrambles from the 2023 Kaggle Santa Challenge dataset with an average solution length of 46.51 - which is below 48 (a conjectured 4$\times$4$\times$4 Rubik's cube diameter~\cite{hirata2024probabilistic}). Finally, an ensemble of 69 agents beat each best solutions for the 5$\times$5$\times$5 Rubik's cube submitted to the 2023 Kaggle Santa Challenge, shortening the average solution rate among all the datasets on more than 4.4 units in QTM metrics (ours: 92.16, Santa: 96.58). It is worth emphasizing that the solutions that were obtained outperformed the Santa results on average and in every single case.

The efficiency of our approach is driven not only by the large number of agents but also by the efficiency of each single node. We performed an additional test to prove this statement and compared it with EfficentCube in terms of average computation time while running on the same hardware. The training procedure for EfficentCube took 86 hours 25 minutes, while the model for our solution was trained in 4 hours 40 minutes. Then, both solutions were used to solve all the scrambles from the DeepCubeA dataset (see results No.9 and 10 in Table~\ref{tbl_best}). Our solution provided slightly better results using the same beam width of $2^{18}$. At the same time, EfficientCube required 287.78~s on average to solve a single scramble, while our solution required 10.91~s, which is $\approx26$ times faster.

%% file: parts/discussion.tex
The paper proposes a machine learning-based approach to the pathfinding problem on large graphs. Experimental studies demonstrate that it is more efficient than state-of-the-art solutions in terms of average solution length, optimality, and computational performance.

The key parts of the approach are multi-agency, neural networks predicting diffusion distance and beam search. Deeper neural networks better approximate the graph of the large Rubik's cubes, though, for the 3$\times$3$\times$3 case, even a single-layer network provides excellent results. At the same time, the effect of enlarging the training set is limited: the trainset above 8196M examples for the tested models has no practical reason, which allowed us to avoid additional time spent during the training.
Conversely, raising the beam width effectively lowers the solution length and increases optimality.

The complete set of the proposed solutions allowed the creation of the multi-agent pathfinder, which managed to beat all the ML-based competitors: an agent equipped with single-layer MLP solved all the DeepCubeA dataset with 90.4\% of optimality significantly enhancing results of the most advanced state of the art solutions: DeepCubeA and EfficientCube. 26 agents equipped with 10-layer ResMLP models managed to solve all scrambles from DeepCubeA dataset with 97.6\% optimality, which is the best result ever achieved by any ML solution. Single-agent solutions implemented using our approach and 10-layer ResMLP beat all the best results corresponding to 3$\times$3$\times$3 and 4$\times$4$\times$4 Rubik's cubes submitted on the 2023 Santa Challenge. At the same time, six agents managed to solve all the 4$\times$4$\times$4 cube's scrambles from the 2023 Santa Challenge dataset with an average solution length below 48 (a 4$\times$4$\times$4 Rubik's cube diameter predicted in \cite{hirata2024probabilistic}). Finally, a composition of 69 agents beat all the best solutions for the 5$\times$5$\times$5 Rubik's cube submitted to the 2023 Santa Challenge, shortening the average solution rate among all the datasets on more than 4.4 units in QTM metrics.

The method's scope is quite wide and can be applied to various planning tasks in their graph pathfinding reformulation. In future works, we plan to explore its applications to mathematical, bioinformatic, and programming tasks.

%% file: parts/methods-I.tex
\subsection{\label{sec_cg_cube}Cayley graphs and Rubik's cubes}

Moves of Rubik's cube can be described by permutations (e.g., Chapter 5~\cite{mulholland2016permutation}, or  Kaggle notebook "Visualize allowed moves"\footnote{\url{https://www.kaggle.com/code/marksix/visualize-allowed-moves}}). Taking all the positions as nodes and connecting them by edges, which differ by single moves, one obtains a Cayley-type (Schreier) graph for Rubik's cube. Solving the puzzle is equivalent to finding a path on the graph between nodes representing the Rubik's cube's scramble initial and solved state.

%% file: parts/methods-II.tex
\subsection{\label{sec_rw_ts}Random walks and train set generation}

The training set is generated by scrambling (i.e., applying random moves) the selected solved state and creating a set of pairs $(v,k)$, where $k$ is a number of scrambles, and  $v$ is a vector describing the node obtained after $k$ steps. In other words, we consider random walks on the graph. The main parameters are $K_{\text{max}}$ and $K$, where $K_{\text{max}}$ is a maximal number of scrambles (length of random walk trajectory), while $K\cdot K_{\text{max}}$ is a number of nodes to generate.

In the current research, we used so-called non-backtracking random walks \cite{alon2007non}, that forbid scrambling to the state of the previous step. A PyTorch-optimized implementation of train set generation can be found in \textit{trainer.py} in the code attached to this paper.

Current research does not investigate the influence of $K_{\text{max}}$ on the solver's performance. We used $K_{\text{max}}=26$ for solvers targeted on 3$\times$3$\times$3 cubes, $K_{\text{max}}=45$ -- for 4$\times$4$\times$4 cubes, and $K_{\text{max}}=65$ for 5$\times$5$\times$5 cubes.

%% file: parts/tab2.tex
\begin{table*}
\caption{The parameters of neural networks used in current research}\label{tbl_neural_net}
\begin{tabular*}{\textwidth}{@{\extracolsep\fill}ccccccccc}
\toprule
\textbf{No.} & \textbf{Cube} & \textbf{Metric}  & \textbf{Layers} & $\boldsymbol{N}_1$ & $\boldsymbol{N}_2$ & $\boldsymbol{N}_r$ & $\boldsymbol{P}$ & \textbf{Result No.} \\
\midrule
1 & 3$\times$3$\times$3 & QTM & 1 & 3050 & 0 & 0 & 1M & -- \\
2 & 3$\times$3$\times$3 & QTM & 2 & 850 & 850 & 0 & 1M & --\\
3 & 3$\times$3$\times$3 & QTM & 6 & 800 & 340 & 2 & 1M & --\\
4 & 3$\times$3$\times$3 & QTM & 10 & 430 & 300 & 4 & 1M & --\\
\midrule
5 & 3$\times$3$\times$3 & QTM & 1 & 12196 & 0 & 0 & 4M & --\\
6 & 3$\times$3$\times$3 & QTM & 2 & 1841  & 1841 & 0 & 4M & --\\
7 & 3$\times$3$\times$3 & QTM & 6 & 2000 & 697 & 2 & 4M & --\\
8 & 3$\times$3$\times$3 & QTM & 10 & 700 & 643 & 4 & 4M & 10, 12, 13\\
\midrule
9 & 4$\times$4$\times$4 & UQTM & 2 & 750 & 750 & 0 & 1M & --\\
10 & 4$\times$4$\times$4 & UQTM & 4 & 530 & 470 & 1 & 1M & --\\
11 & 4$\times$4$\times$4 & UQTM & 6 & 720 & 300 & 2 & 1M & --\\
12 & 4$\times$4$\times$4 & UQTM & 10 & 500 & 266 & 4 & 1M & --\\
\midrule
13 & 4$\times$4$\times$4 & UQTM & 2 & 1730  & 1730 & 0 & 4M & --\\
14 & 4$\times$4$\times$4 & UQTM & 6 & 1180 & 1024 & 1 & 4M & --\\
15 & 4$\times$4$\times$4 & UQTM & 6 & 2000 & 628 & 2 & 4M & --\\
16 & 4$\times$4$\times$4 & UQTM & 10 & 1010 & 592 & 4 & 4M & 18, 19\\
\midrule
17 & 4$\times$4$\times$4 & UQTM & 6 & 2000 & 1126 & 2 & 8M & --\\
18 & 4$\times$4$\times$4 & UQTM & 10 & 1540 & 850 & 4 & 8M & --\\
\midrule
19 & 4$\times$4$\times$4 & UQTM & 6 & 5000 & 1369 & 2 & 16M & --\\
20 & 4$\times$4$\times$4 & UQTM & 10 & 5000 & 1062 & 4 & 16M & --\\
\midrule
21 & 2x2x2 & QTM & 1 & 1024 & 0 & 0 & 0.15M & 3\\
22 & 2x2x2 & QTM & 10 & 430 & 300 & 4 & 0.92M & 4\\
23 & 3$\times$3$\times$3 & QTM & 1 & 1024 & 0 & 0 & 0.34M & 11\\
24 & 3$\times$3$\times$3 & UQTM & 10 & 700 & 643 & 4 & 4M & 15\\
25 & 5$\times$5$\times$5 & UQTM & 10 & 1008 & 560 & 4 & 4M & 22\\
\bottomrule
\end{tabular*}

\end{table*}

%% file: parts/methods-III.tex
\subsection{\label{sec_train_procedure}Neural Network and Training procedure}

In this study, we used ResMLP, a generalized form of multilayer perceptrons as described in \cite{agostinelli2019solving,takano2023selfsupervision}. 
Details of the architecture can be found in Figure~\ref{fig_solution_struct}b.
The PyTorch implementation of ResMLP is available in \textit{model.py} in the code attached to this paper.

The training procedure was performed using the Adam optimizer with a fixed learning rate of 0.001 and mean squared error as the loss function. 
A new dataset of 1M examples was generated before each training epoch. 
All models were pre-trained and remained unchanged during puzzle-solving. 
Training was conducted using 32-bit floating point precision, while inference used 16-bit floating point numbers to enhance computational efficiency. 
The PyTorch implementation of the training procedure is available in \textit{trainer.py} in the code attached to this paper.

%% file: parts/methods-IV.tex
\subsection{\label{sec_search}Beam-search}
Beam search is a simple but effective search procedure used for various optimization tasks~\cite{hale2018finding, huang2019linearfold, scheidl2018word} as well as to improve outputs of the modern transformer-based language models~\cite{freitag2017beam, pryzant2023automatic, musolesi2024creative}. It has been used in EfficientCube~\cite{takano2023selfsupervision} and by many participants of the Kaggle Challenge~\cite{santa-2023}. We implemented a modified version of traditional beam search, which uses hash functions to remove duplicates, reducing the computation complexity of the pathfinder. Finally, in all the experiments, the scramble was considered unsolved if the path to the solved state was not found in 200 beam search steps. Additionally, the algorithm stops if the beam vector contains only already visited graph nodes. A PyTorch-optimized implementation of the beam search can be found in \textit{searcher.py} in the code attached to this paper.

%% file: parts/tab3.tex
\begin{table*}

\caption{Average solution length depending from trainset size $T$ and beam width $W$ for tested with puzzle win probability above 0.5.}\label{tbl_beam_vs_train}

\begin{tabular*}{\textwidth}{@{\extracolsep\fill}c|cc|cc|cc}
\toprule%
\textbf{\textit{T}} & \multicolumn{2}{c|}{\textbf{4M}} & \multicolumn{2}{c|}{\textbf{8M}} & \multicolumn{2}{c}{\textbf{16M}}\\
\midrule
\textbf{\textit{W}} & \textbf{Win prob.} & \textbf{Avg. sol.} & \textbf{Win prob.} & \textbf{Avg. sol.} & \textbf{Win prob.} & \textbf{Avg. sol.} \\
\midrule
\multicolumn{7}{c}{ \textbf{3$\times$3$\times$3 Rubik's solver results with win probability above 0.5}}\\
\midrule
$2^{12}$ & 1.00 & 22.15 & 1.00 & 22.05 & 1.00 & 22.3\\
$2^{14}$ & 1.00 & 21.55 & 1.00 & 21.55 & 1.00 & 21.6\\
$2^{16}$ & 1.00 & 21.25 & 1.00 & 21.15 & 1.00 & 21.2\\
$2^{18}$ & 1.00 & 21.15 & 1.00 & 20.95 & 1.00 & 21\\
$2^{20}$ & 1.00 & 20.75 & 1.00 & 20.85 & 1.00 & 20.9\\
$2^{22}$ & 1.00 & 20.65 & 1.00 & 20.65 & 1.00 & 20.7\\
$2^{24}$ & 1.00 & 20.65 & 1.00 & 20.65 & 1.00 & 20.65\\
\midrule
\multicolumn{2}{l}{\textbf{Average solution}} & \textbf{21.16} & & \textbf{21.12} & & \textbf{21.19}\\ 
\midrule
\multicolumn{7}{c}{ \textbf{4$\times$4$\times$4 Rubik's solver results with win probability above 0.5}}\\
\midrule
$2^{16}$ & 0.80 & 61.88 & 0.55 & 60.18 & 0.85 & 64.12\\
$2^{18}$ & 1.00 & 58.1 & 0.85 & 56.7 & 0.95 & 58.47\\
$2^{20}$ & 1.00 & 54.7 & 1.00 & 55.3 & 1.00 & 54.7\\
$2^{22}$ & 1.00 & 51.9 & 1.00 & 52.8 & 1.00 & 52.3\\
$2^{24}$ & 1.00 & 50.4 & 1.00 & 50.6 & 1.00 & 49.6\\
\midrule
\multicolumn{2}{l}{\textbf{Average solution}} & \textbf{55.4} & & \textbf{55.11} & & \textbf{55.84}\\ 
\midrule
\end{tabular*}

\end{table*}

%% file: parts/methods-V.tex
\subsection{\label{sec_exp_design}Experiments design}

All the experiments were conducted using software attached to this paper. The experiments targeting analysis of trainset size's influence on the solver's performance included solving 20 scrambles of both 3$\times$3$\times$3 and 4$\times$4$\times$4 Rubik's cubes using different models as beam search heuristics. For this experiment, we prepared 20 models, whose parameters are demonstrated in the first 20 rows of Table~\ref{tbl_neural_net}. Each model was trained during 16384 epochs. The snapshots of the model parameters were saved after 16, 64, 256, 1024, 4096, and 16384 epochs. Then, each model snapshot was integrated as a heuristic into beam search with $W=2^{18}$, which was used to solve the first 20 scrambles from the dataset. DeepCubeA dataset~\cite{agostinelli2019solving} was used for 3$\times$3$\times$3 Rubik's cube, and the 2023 Kagle Santa Challenge~\cite{santa-2023} dataset was used for 4$\times$4$\times$4 puzzle. The results achieved by each solver configuration on the corresponding dataset were averaged and analyzed as experimental results. Unsolved scrambles were excluded from consideration in this experiment.

The first experiment's results are demonstrated in Figure~\ref{fig_param_vs_len}a. Single layer MLPs for 4$\times$4$\times$4 Rubik's cube are not presented in Table~\ref{tbl_neural_net} as during preliminary research solvers equipped with this type of model did not manage to solve any scramble before reaching the 200 steps limit.

During the second experiment, we used only 10 layer models with size 4M (models No.8 and 16 from Table~\ref{tbl_neural_net}). The first experiment's results did not show a significant effect of increasing $T$ (train size) from 4B to 16B. Thus, the second experiment used finer granularity with 4B, 8B, and 16B train sizes to select the appropriate size more precisely. Each snapshot of the models trained with the mentioned above train sets was integrated into solvers with $W$ of $2^{12}$, $ 2^{14}$, $2^{16}$, $2^{18}$, $2^{20}$,  $2^{22}$, and $2^{24}$. Then, these solvers were used to unscramble the first 20 puzzles from the same dataset used in the first experiment. The results achieved by each solver configuration on the corresponding dataset were averaged. Unsolved scrambles were excluded from consideration in this experiment.

The results of the second experiment are demonstrated in Figure~\ref{fig_param_vs_len}b. A deeper analysis of the experimental results shows that if we consider only $W$ values that give a puzzle winning probability above 50\%, the agent with the model trained on 8B states has a slightly better average solution length than the competitors~(Table~\ref{tbl_beam_vs_train}). Thus, for the rest of the experiments, we used the 8B train set as a compromise between solver performance and training time.

The third experiment analyzed the influence of the number of agents $A$ on the solver's efficiency. We used models No.8, 16, and 25 for this experiment from Table~\ref{tbl_neural_net}. We trained each of these models multiple times during 8192 epochs. Then, each model was integrated into a dedicated agent. Due to computation limitations, we run only two agents in parallel at the same time, assuming that with more available GPU instances, it will be possible to compute all of them simultaneously. Finally, the total number of pretrained agents for 3$\times$3$\times$3 was 26, 4$\times$4$\times$4 -- 29, and 5$\times$5$\times$5 -- 69. As in previous experiments, the agents aimed to solve 3$\times$3$\times$3 cubes were tested on the scrambles DeepCubeA dataset, while the rest were verified using the 2023 Kagle Santa Challenge dataset. Due to the large size of the DeepCubeA dataset, in the third experiment, we used a subset of 69 3$\times$3$\times$3 scrambles, which were considered most difficult during preliminary research. The results of the experiment are shown in Figure~\ref{fig_multi_agent}.

The authors of~\cite{agostinelli2019solving}, along with the well-known DeepCubeA dataset, were using DeepCubeA$_h$ set containing the scrambles that are furthest away from the goal state, assuming these scrambles are more challenging to solve. At the same time, original DeepCubeA solutions were robustly and optimally solving them. During the current research, we found another subset of the DeepCubeA dataset containing 16 scrambles, which were not solved optimally during the experimental studies. We believe that a significantly rising number of agents will lead to finding solutions to all of them. However, the first element of this subset~(scramble No.17 from original DeepCubeA) was never optimally solved in any of our attempts, even during preliminary research and experiments not covered by this paper. We believe that analyzing the scrambles in this subset will help us understand why they are so hard to solve compared to the rest of the DeepCubeA data. Finally, this understanding will lead to the development of new, more efficient ML methods.\footnote{A possible explanation is that the number of optimal solution paths for such cubes is lower than average or equal to one, making these paths more difficult to find.} Thus, we decided to publish these 16 scrambles as a self-contained dataset accompanied by this paper.

The experimental results listed in Table~\ref{tbl_best} were achieved by solving all the scrambles from the corresponding dataset defined in the third column of Table~\ref{tbl_best} using the proposed solver. The key solver parameters are listed in the fourth column. The last column of Table~\ref{tbl_neural_net} demonstrates for which results from Table~\ref{tbl_best} each model was used.

The last experiment conducted in the current research was aimed to compare computational efficiency with the EfficientCube~\cite{takano2023selfsupervision} (a state-of-the-art solution claimed by its author to have better efficiency than DeepCubeA). For the fairness of comparison, we set up a virtual machine with the following resources: AMD EPYC 7513 32-core Processor running at 2.6GHz; 240\,GB RAM; 250\,GB NFS file storage; a single dedicated GPU NVIDIA A100-SXM 80GB. The virtual machine ran Red Hat Enterprise Linux version 8.7 and CUDA version 11.8. The latest version~(March 10th, 2024) of the EfficientCube was downloaded from the official GitHub repository\footnote{\url{https://github.com/kyo-takano/efficientcube}} and configured according to the author's instructions to reproduce the results from~\cite{takano2023selfsupervision}. Our solution was installed on the same virtual machine and configured with the same beam width of $2^{18}$. First, we sequentially trained a model for each solution and measured the time required for these procedures. Then, we tested both solvers on all the scrambles from the DeepCubeA dataset. We recorded the solving time for each scramble and then averaged it among the whole dataset. Finally, we compared training time and average solving time between EfficientCube and our solution. During this experiment, only one solution was running on the virtual machine at the same time.

%% file: parts/acknowledgments.tex
We express our gratitude to Graviton for providing servers for research in the field of AI. 

A.C. is deeply grateful to M. Douglas for his interest in this work, engaging discussions, and invitation to present preliminary results at the Harvard CMSA program on "Mathematics and Machine Learning" in Fall 2024, to M. Gromov, S. Nechaev, and V. Rubtsov for their invitation to make a talk at "Representations, Probability, and Beyond: A Journey into Anatoly Vershik’s World" at IHES, as well as for stimulating discussions. A.C. is grateful to J. Mitchel for involving  into the Kaggle Santa 2023 challenge, from which this project originated and to M. Kontsevich, Y. Soibelman, A. Soibelman, S. Gukov, A. Hayat, T. Smirnova-Nagnibeda,  D. Osipov, V. Kleptsyn, G. Olshanskii, A. Ershler, J. Ellenberg, G. Williamson, A. Sutherland,  Y. Fregier, P.A. Melies, I. Vlassopoulos, F. Khafizov, A. Zinovyev,  H. Isambert for the discussions, interest and comments. 

K.K. is grateful to T. Ruzaikin, A. Fokin, and M. Goloshchapov for fruitful discussions.

We are deeply grateful to the many colleagues who have contributed to the CayleyPy project at various stages of its development, including: S.Fironov, A.Lukyanenko, A. Abramov, A. Ogurtsov, A. Trepetsky, A. Dolgorukova, S. Lytkin, S. Ermilov, L. Grunvald, A. Eliseev, G. Annikov, M. Evseev, F. Petrov, N. Narynbaev, S. Nikolenko, S. Krymskii, R. Turtayev, S. Kovalev, N. Rokotyan, G. Verbyi, L. Shishina, A. Korolkova, D. Mamaeva, M. Urakov, A. Kuchin, V. Nelin, B. Bulatov,  F. Faizullin, A. Aparnev, O. Nikitina, A. Titarenko, U. Kniaziuk, D. Naumov, A. Krasnyi, S. Botman, R. Vinogradov, D. Gorodkov, I. Gaiur, I. Kiselev, A. Rozanov, K. Yakovlev, V. Shitov, E. Durymanov, A. Kostin, R. Magdiev, M. Krinitskiy, P. Snopov.

%% file: parts/availability.tex
\section{Data availability}
The weights and datasets used in the experimental studies are openly available on Zenodo~\cite{chervov_dataset}.

The dataset containing a subset of the 16 most difficult scrambles from the DeepCubeA dataset that were not solved optimally with our approach is available at~\cite{chervov_hard}.

\phantom{42}

\section{Code availability}

The source code used to perform experimental studies in current research is available on GitHub~\footnote{\url{https://github.com/k1242/cayleypy-cube}}.

Notebooks related to the CayleyPy project development are available on Kaggle~\footnote{\url{https://www.kaggle.com/datasets/alexandervc/growth-in-finite-groups}}, ~\footnote{\url{https://www.kaggle.com/competitions/lrx-oeis-a-186783-brainstorm-math-conjecture/code}}.